\documentclass[10pt,twocolumn,letterpaper]{article}

\usepackage{cvpr}
\usepackage{times}
\usepackage{epsfig}
\usepackage{graphicx}
\usepackage{amsmath}
\usepackage{amssymb}
\usepackage{microtype}
\usepackage{booktabs}
\usepackage{tabularx}
\usepackage{xcolor}

\makeatletter
\newcommand{\printfnsymbol}[1]{%
  \textsuperscript{\@fnsymbol{#1}}%
}
\makeatother
\usepackage{epsfig}
\usepackage[title]{appendix}
\usepackage[caption=false]{subfig}
\usepackage{tabulary,multirow,overpic,xcolor}
\newcolumntype{x}[1]{>{\centering\arraybackslash}p{#1pt}}

\newlength\savewidth\newcommand\shline{\noalign{\global\savewidth\arrayrulewidth
  \global\arrayrulewidth 1pt}\hline\noalign{\global\arrayrulewidth\savewidth}}
\newcommand{\tablestyle}[2]{\setlength{\tabcolsep}{#1}\renewcommand{\arraystretch}{#2}\centering\footnotesize}

\usepackage[pagebackref=true,breaklinks=true,letterpaper=true,colorlinks,bookmarks=false]{hyperref}

\cvprfinalcopy 

\def\cvprPaperID{1583} 

\ifcvprfinal\fi
\begin{document}

\title{Actor-Context-Actor Relation Network for Spatio-Temporal Action Localization}

\author{%
Junting Pan$^{1}$\thanks{Equal contribution} \quad\quad\quad Siyu Chen$^{4*}$ \quad\quad\quad Mike Zheng Shou$^2$ \\
\ \quad Yu Liu$^4$ \quad\quad\quad\quad Jing Shao$^4$ \ \quad\quad\quad\quad Hongsheng Li$^{1,3}$ \vspace{.5em} \\
$^1$CUHK-SenseTime Joint Lab, The Chinese University of Hong Kong  \quad $^2$Columbia University  \\
$^3$School of CST, Xidian University \quad $^4$ SenseTime Research
\vspace{-.5em}
}

\maketitle

\begin{abstract}
Localizing persons and recognizing their actions from videos is a challenging task towards high-level video understanding. 
Recent advances have been achieved by modeling direct pairwise relations between entities. 
%
%
In this paper, we take one step further, not only model direct relations between pairs but also take into account indirect higher-order relations established upon multiple elements.
We propose to explicitly model the \textbf{Actor-Context-Actor Relation}, which is the relation between two actors based on their interactions with the context.
To this end, we design an Actor-Context-Actor Relation Network (ACAR-Net) which builds upon a novel \textit{High-order Relation Reasoning Operator} and an \textit{Actor-Context Feature Bank} to enable indirect relation reasoning for spatio-temporal action localization. 
%
%
Experiments on AVA and UCF101-24 datasets show the advantages of modeling actor-context-actor relations, and visualization of attention maps further verifies that our model is capable of finding relevant higher-order relations to support action detection.
%
Notably, our method ranks first in the AVA-Kinetics action localization task of ActivityNet Challenge 2020, outperforming other entries by a significant margin (+6.71 mAP).
The code is available online.\footnote{\url{https://github.com/Siyu-C/ACAR-Net}}
\end{abstract}

 
\section{Introduction}
\label{sec:intro}

\begin{figure}[!t]
\centering
\includegraphics[width=\linewidth]{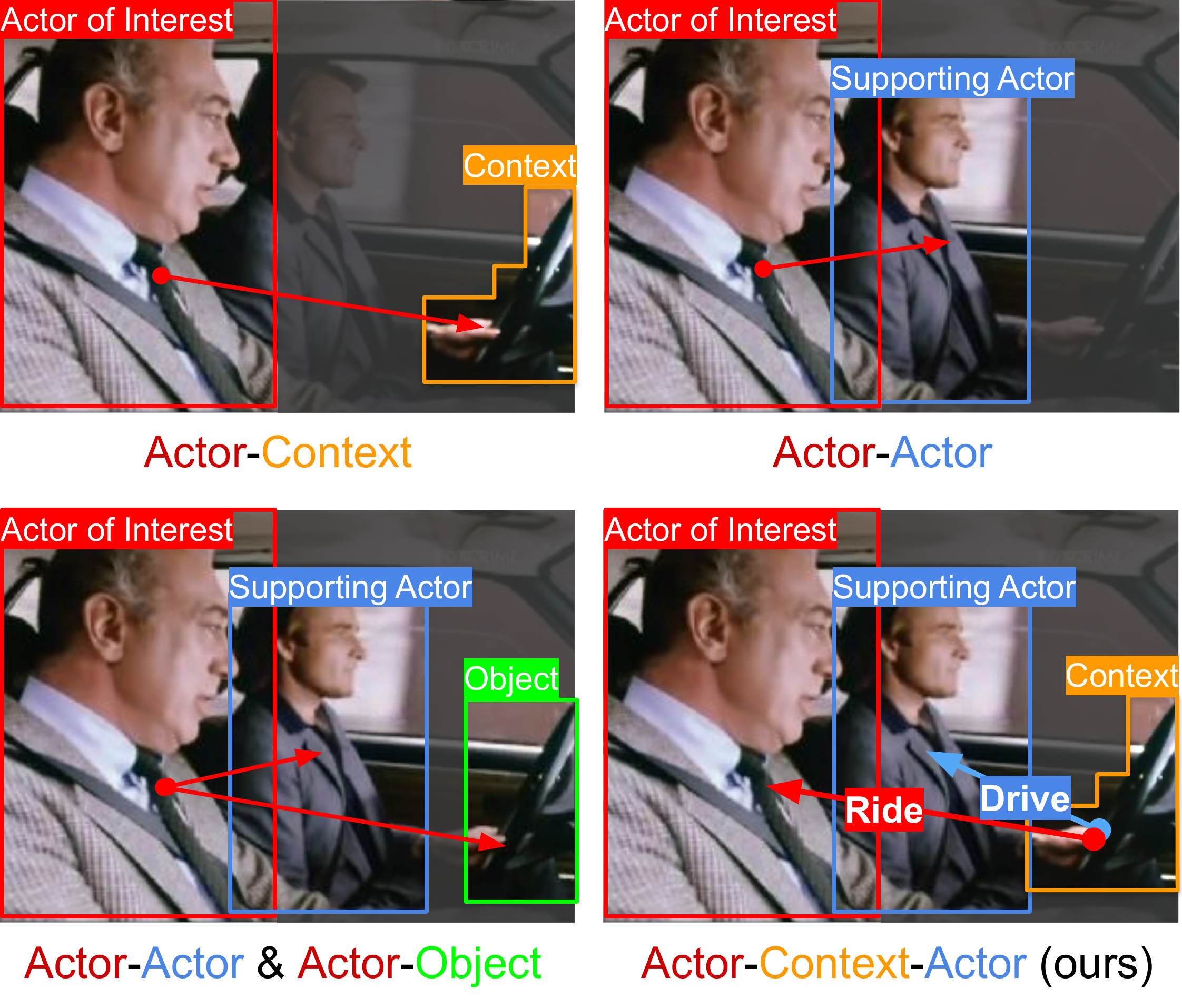}
\caption{We contrast our Actor-Context-Actor relation modeling with existing relation reasoning approaches for action localization.
Reasoning relations between pairs of entities may not always be sufficient for correctly predicting the action labels of all individuals. Our method not only reasons relations between actors, but also models connections between different actor-context relations. As an illustration, the relation between the blue actor and the steering wheel (drive) serves as a crucial clue for recognizing the action being performed by the red actor (ride). \label{fig:figure1}
}
\vspace{-4mm}
\end{figure}

Spatio-temporal action localization, which requires localizing persons and recognizing their actions from videos, is an important task that has drawn increasing attention in recent years \cite{gu2018ava, girdhar2019video, feichtenhofer2019slowfast, wu2019long, sun2018actor, zhang2019structured, xie2018rethinking, yang2019step, ulutan2020actor, peng2016multi, ye2019discovering, hou2017tube, kalogeiton2017action}. Unlike object detection which can be accomplished solely by observing visual appearances, activity recognition usually demands for reasoning about the actors' interactions with the surrounding context, including environments, other people and objects. Take Fig.~\ref{fig:figure1} as an example. To recognize the action ``ride'' of the person in the red bounding box, we need to observe that he is inside a car, and there is a driver next to him.
Therefore, most recent progress in spatio-temporal action detection has been driven by the success of relation modeling. These approaches focus on modeling relationships in terms of pairwise interactions between entities. 

However, it is not always the case that relations between elements can be formulated in terms of pairs; often, higher-order relations provide crucial clues for accurate action detection. In Fig.~\ref{fig:figure1}, it is difficult to infer the action of the red actor given only its relation with the blue actor, or only with the scene context (steering wheel). Instead, in order to identify that the red actor performs the action ``ride", one has to reason over the interaction between the blue actor and the context (drive). In other words,  it is necessary to capture the implicit \textit{second-order} relation between the two actors based on their respective \textit{first-order} relations with the context.

There were previous works that employ Graph Neural Networks (GNNs) to implicitly model higher-order interactions between actors and contextual objects~\cite{wang2018videos, zhang2019structured, tang2020asynchronous, zellers2018neural, gao2020drg}. However, in these approaches, an extra pre-trained object detector is required, and only located objects are used as context. Since bounding-box annotations of objects in spatio-temporal action localization datasets are generally not provided, the pre-trained object detector is limited to its original object categories and may easily miss various objects in the scenes. In addition, the higher-order relations in these methods are limited to be inferred solely from contextual objects, which might miss important environmental or background cues for action classification.

To tackle the above issues, we propose an Actor-Context-Actor Relation Network (ACAR-Net) 
which focuses on modeling second-order relations in the form of Actor-Context-Actor relation. It deduces indirect relations between multiple actors and the context for action localization. The ACAR-Net takes both actor and context features as inputs. We define actor features as the features pooled from the actor regions of interest, while for context features, we directly use spatio-temporal grid feature maps from our backbone network. 
The context that we adopt does not rely on any extra object detector with predefined categories, thus making our overall design much simpler and flexible.
Moreover, grid feature maps are capable of representing scene elements of various levels (\eg instance level and part level) and types (\eg background, objects and object parts), which is useful for fine-grained action discrimination. 
The proposed ACAR-Net first encodes first-order actor-context relations, and then applies a \textit{High-Order Relation Reasoning Operator} to model interactions between the first-order relations. The High-Order Relation Reasoning Operator is fully convolutional and operates on first order relational features maps without losing spatial layouts 
. For supporting actor-context-actor relation reasoning between actors and context at different time periods, we build an \textit{Actor-Context Feature Bank}, which contains actor-context relations from different time steps across the whole video.

We conduct extensive experiments on the challenging Atomic Visual Actions (AVA) dataset~\cite{gu2018ava, li2020ava} as well as the UCF101-24 dataset~\cite{soomro2012ucf101} for spatio-temporal action localization. 
Our proposed ACAR-Net leads to significant improvements on recognizing human-object and human-human interactions. Qualitative visualization shows that our method learns to attend contextual regions that are relevant to the action of interest.

Our contributions are summarized as the three-fold: 
\begin{itemize}
\item We propose to model actor-context-actor relations for spatio-temporal action localization. Such relations are mostly ignored by previous methods but crucial for achieving accurate action localization.
\item We propose a novel Actor-Context-Actor Relation Network for improving spatio-temporal action localization by explicitly reasoning about higher-order relations between actors and the context.
\item We achieve state-of-the-art performances with significant margins on the AVA and UCF101-24 datasets. At the time of submission, our method ranks first on the ActivityNet leaderboard \cite{caba2015activitynet}.
\end{itemize}

\section{Related Work}
\label{sec:RelatedWork}

{\flushleft \bf Action Recognition.}
Research works on action recognition generally fall into three categories: action classification, temporal localization and spatio-temporal localization. Early works mainly focus on classifying a short video clip into an action class. 3D-CNN \cite{tran2015learning, carreira2019short, feichtenhofer2019slowfast}, two-stream network \cite{simonyan2014two, wang2016temporal,feichtenhofer2016convolutional} and 2D-CNN \cite{yue2015beyond, donahue2015long, lin2019tsm} are the three dominant network architectures adopted for this task. While progress has been made for short trimmed video classification, the main research stream also moves forward to understand long untrimmed videos, which requires not only to recognize the category of each action instance but also to locate its start and end times. A handful of works \cite{shou2016temporal, xu2017r, Chao_2018_CVPR, zhao2017temporal} consider this problem as a detection problem in 1D temporal dimension by extending object detection frameworks.
{\flushleft \bf Spatio-Temporal Action Localization.}
Recently, the problem of spatio-temporal action localization has drawn considerable attention from the research community, and datasets (such as AVA \cite{gu2018ava, li2020ava}) with atomic actions of all actors in the video being continuously annotated are introduced. It defines the action detection problem into a finer level, since the action instances need to be localized in both space and time. Typical approaches used by early works apply R-CNN detectors on 3D-CNN features \cite{gu2018ava, girdhar2018better, yang2019step, xiao2020audiovisual, li2018recurrent}. Wu \etal \cite{wu2020context} show that actor features obtained by running 3D-CNN backbone on top of the cropped and resized actor region from the original video preserve better spatial details than RoI-pooled actor features. Nevertheless, it has the limitation that computational costs and inference time almost increase linearly with the number of actors. Several more recent works have exploited graph-structured networks to leverage contextual information \cite{sun2018actor, girdhar2019video, zhang2019structured, tang2020asynchronous, ulutan2020actor, tomei2019stage}. 

{\flushleft \bf Relational Reasoning for Video Understanding.}
Relational reasoning has been studied in the domain of video understanding \cite{wang2018non, wang2018videos, zhou2017temporal, sun2019relational, zhang2019structured, sun2018actor, wu2019long, ji2020action, materzynska2020something, tang2020asynchronous}. 
This is natural because recognizing the action of an actor depends on its relationships with other actors and objects.
Zhou \etal \cite{zhou2017temporal} extend Relation Network \cite{santoro2017simple} for modeling relations between video frames over time. Non-local Networks \cite{wang2018non} leverage self-attention mechanisms to capture long range dependencies between different entities. Wang \etal \cite{wang2018videos} show that representing videos with Space-time Region Graph improves action classification accuracy. 
In the context of spatio-temporal localization, there are many traditional approaches that are dedicated to capturing spatio-temporal relationships in videos \cite{zhang2013modeling, morariu2011multi, swears2014complex, intille1999framework}. For deep neural networks based methods, Sun \etal \cite{sun2018actor} propose Actor-Centric Relation Network that learns to aggregate actor and scene features. 
Girdhar \etal \cite{girdhar2019video} re-purpose the Transformer network \cite{vaswani2017attention} for encoding pairwise relationships between every two actor proposals. 
%
%
Concurrently, Wu \etal \cite{wu2019long} use long-term feature banks (LFB) to provide temporal supportive information up to 60s for computing long range interaction between actors.
Zhang \etal \cite{zhang2019structured} propose to explicity model interactions between actors and objects.
However, their approach focuses on modeling actor-object and actor-actor relations separately. When deducing the action of a person, the interactions of other persons with contextual objects are ignored. In other words, they do not explicitly model the actor-context-actor relations. In contrast, our method emphasizes modeling those higher-order relations. Perhaps the most similar work to ours is \cite{tang2020asynchronous}, which aggregates multiple types of interactions with stacked units akin to Transformer Networks \cite{vaswani2017attention}. Nonetheless, while this approach also supports actor-context-actor interactions, it treats object detection results as context, which requires extra pre-trained object detectors with fixed object categories and ignores other important types of contexts (such as background, objects not in the predefined categories, and specific parts of some objects).

%



\section{Method}
\label{sec:Method}
In this section, we provide detailed descriptions of our proposed Actor-Context-Actor Relation Network (ACAR-Net). 
Our ACAR-Net aims at effectively modeling and utilizing higher-order relations built upon the basic actor-actor and actor-context relations for achieving more accurate action localization.


\begin{figure*}[t]
\centering
\includegraphics[width=\textwidth]{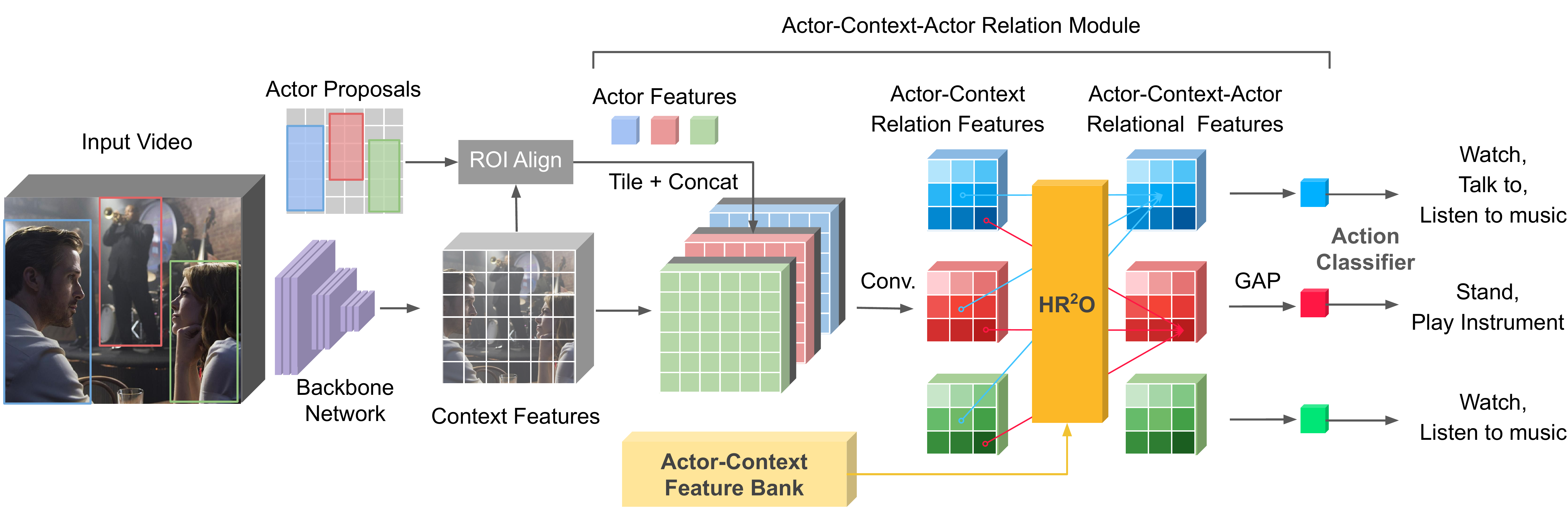}
\caption{ \textbf{Action Detection Framework.} Videos are processed with a Backbone Network to produce spatio-temporal context features. For each actor proposal (person bounding box), we extract actor features from the context features by RoIAlign. Given the actor and context features, the ACAR-Net computes second-order relation between every two actors based on their interaction with the context. 
\label{fig:pipeline}
}
\vspace{-5mm}
\end{figure*}

\subsection{Overall Framework}
We first introduce our overall framework for action localization, where the proposed actor-context-actor relation (ACAR) modeling is the key module. The framework is designed to detect all persons in an input video clip ($\sim$2s 
in our experiments) and estimate their action labels. 
As shown in Fig.~\ref{fig:pipeline}, following state-of-the-art methods~\cite{wu2019long,feichtenhofer2019slowfast,xia2019three}, the framework is built based on an {off-the-shelf person detector} (\textit{e.g.} Faster R-CNN~\cite{ren2015faster}) and a {video backbone network} (\textit{e.g.} I3D~\cite{carreira2017quo}). Person and context features are then processed by the proposed ACAR module with a long-term \textit{Actor-Context Feature Bank} for final action prediction. 

In details, the person (actor) detector operates on the center frame (\ie key frame) of the input clip and obtains $N$ detected actors. The detected boxes are duplicated to neighboring frames of the key frame in the clip.
In the meantime, the backbone network extracts a spatio-temporal feature volume from the input video clip. We perform average pooling along the temporal dimension to save follow-up computational cost, which results in a feature map $X \in \mathbb{R}^{C\times H\times W}$, and $C, H, W$ are channel, height and width respectively. We apply RoIAlign~\cite{he2017mask} ($7\times 7$ spatial output) followed by spatial max pooling to the $N$ actor features, producing a series of $N$ actor features, $A^1, A^2, \dots, A^N \in \mathbb{R}^C$, each of which describes the spatio-temporal appearance and motion of one Region of Interest (RoI).


The proposed Actor-Context-Actor Relation (ACAR) module is illustrated on the right side of Fig.~\ref{fig:pipeline}. This module takes the aforementioned video feature map $X$ and RoI features $\{A^i\}_{i=1}^N$ as inputs, and outputs the final action predictions after relation reasoning. The ACAR module has two main operations. (1) It first encodes  first-order actor-context relations between actors and spatial locations of the spatio-temporal context. Based on the actor-context relations, we further integrate a \textit{High-order Relation Reasoning Operator} (HR$^2$O) for modeling the interactions between pairs of first-order relations, which are indirect relations mostly ignored by previous methods.
(2) Our reasoning operation is extended with an {Actor-Context Feature Bank} (ACFB). The bank contains actor-context relations at different time stamps, and can provide more complete spatio-temporal context than the existing long-term feature bank \cite{wu2019long} which only consists of features of actors. We will elaborate the two parts in the following sections. Notably, our high-order relation reasoning only requires action labels as supervision. 

\subsection{Actor-Context-Actor Relation Modeling}
\label{sec:hro}

\noindent \textbf{First-order actor-context relation encoding.} We adopt the Actor-Centric Relation Network (ACRN) \cite{sun2018actor} as a module for encoding the first-order actor-context relations by combining RoI features $A^1,$ $\dots, A^N$ with the context feature $X$. More specifically, it replicates and concatenates each actor feature $A^i \in \mathbb{R}^{C}$ to all $H\times W$ spatial locations of the context feature $X \in \mathbb{R}^{C \times H \times W}$ to form a series of concatenated feature map $\{\tilde{F}^{i}\}_{i=1}^N \in \mathbb{R}^{2C \times H \times W}$. Actor-context relation features for each actor $i$ can then be encoded by applying convolutions to this concatenated feature map ${\tilde{F}^{i}}$.

\noindent \textbf{High-order relation reasoning.}
%
We now discuss how to compute high-order relations between two actors based on their first-order interactions with the context.
Let $F^{i}_{x,y}$ record the first-order features between the actor $A^i$ and the scene context $X$ at the spatial location $(x,y)$. We propose to model the relationship between first-order actor-context relations, which are high-order relations encoding more informative scene semantics. However, since there are a large number of actor-context relation features, $F^i_{x,y} \in \mathbb{R}^{C \times 1 \times 1}$, $i \in \{1,\dots, N\}, x \in [1,H], y \in [1,W]$, the number of their possible pairwise combinations is generally overwhelming. We therefore design 
a \textit{High-order Relation Reasoning Operator} (HR$^2$O) that aims at learning the high-order relations between pairs of actor-context relations at the same spatial location $(x,y)$, \ie,\ $F^i_{x,y}$ and $F^j_{x,y}$. In this way, the proposed relational reasoning operator limits the relation learning to second-order actor-context-actor relations, \ie two actors $i$ and $j$ can be associated via the same spatial context, denoted as $i \leftrightarrow (x,y) \leftrightarrow j$, to help the estimation of their actions. 

Our proposed HR$^2$O takes as input a set of first-order actor-context relation feature maps $\{F^i\}_{i=1}^N$.
%
The operator outputs
$\{H^i\}^{N}_{i=1}=\text{HR}^2\text{O}(\{F^i\}^{N}_{i=1})$ 
that encode second-order actor-context-actor relations for all actors. 
The operator is modeled as stacking several modified non-local blocks~\cite{wang2018non}. 
For each non-local block, 
convolutions are used to convert the input first-order actor-context relation feature maps $F^i$ into query $Q^i$ , key $K^i$ and value $V^i$ embeddings of the same spatial size as $F^i$. All feature maps are of dimension $d=512$ in our implementation. It is worth noting that the use of convolutions is not only useful for aggregating local information but also makes the operator position and order-sensitive. The attention vectors are computed separately at every spatial location, and the Actor-Context-Actor Relation feature $H^i$ is given by the linear combination of all value features $\{V^j\}_{j=1}^N$ according to their corresponding attention weights $Att^{i,j}$. The overall process can be summarized by the following equations,
\begin{equation}
    \vspace{-3mm}
    \begin{aligned}
    {Q^i, K^i, V^i}&=\mathrm{conv2D}(F^i)\\
    {Att}^{i}_{x,y}&= \mathrm{softmax}_j\left(\frac{\left\langle{Q^{i}_{x,y}},{K^{j}_{x,y}}\right\rangle}{\sqrt{d}}\right),\\
    \tilde{H}^{i}_{x,y}&=\sum_{j}Att^{i,j}_{x,y}{V^{j}_{x,y}}.
    \end{aligned}
\end{equation}
Following \cite{wu2019long}, we also add layer normalization and dropout to our modified non-local block,
\begin{equation}
    \begin{aligned}
    {H^{i}}&=\mathrm{Dropout}(\mathrm{Conv2D}(\mathrm{ReLU}(\mathrm{norm}({\tilde{H}^{i}})))), \\
    {F'}^i &= F^i + H^i,
    \end{aligned}
\end{equation}
where $H^i$ and the input actor-context features $F^i$ are fused via residual addition to obtain the actor-context-actor feature ${F'}^i$, which can be further processed by the following non-local block again.

We also exploit another instantiation, which directly obtains second-order actor-context-actor interaction features from actor features $\{A^i\}_{i=1}^N$ and the context feature $X$ by a Relation Network~\cite{santoro2017simple}. More specifically, we obtain the relation feature between actors $A^i$, $A^j$ and context $V_{x,y}$ as 
\begin{equation}
    H^{i,j}_{x,y} = f_\theta([A^i, A^j, V_{x,y} ]),
\end{equation}
where $[\cdot,\cdot,\cdot]$ denotes concatenation along the channel dimension and $f_\theta(\cdot)$ is a stack of two convolutional layers.
The high-order relation of an actor $i$ is calculated as the average of all relation features related to that actor,
\begin{equation}
    H^{i} =  \frac{1}{N} \sum_{j} H^{i,j}_{x,y}.
    \vspace{-3mm}
\end{equation}
It is also fused with the input features to obtain actor-context-actor features via residual addition, \ie ${F'}^i = F^i + H^i$.
This method is computationally expensive when the number of actors $N$ is large, since the number of feature triplets is proportional to $N^2$.

\noindent {\bf Action classifier.} After the actor-context-actor feature maps $\{{F'}^i\}_{i=1}^N$ are obtained for all actors, a final action classifier is introduced as a single fully-connected layer with a non-linearity function to output the confidence scores of each actor belonging to different action classes.




\subsection{Actor-Context Feature Bank}
\label{sec:acfb}

\begin{figure}[t]
\centering
\includegraphics[width=\linewidth]{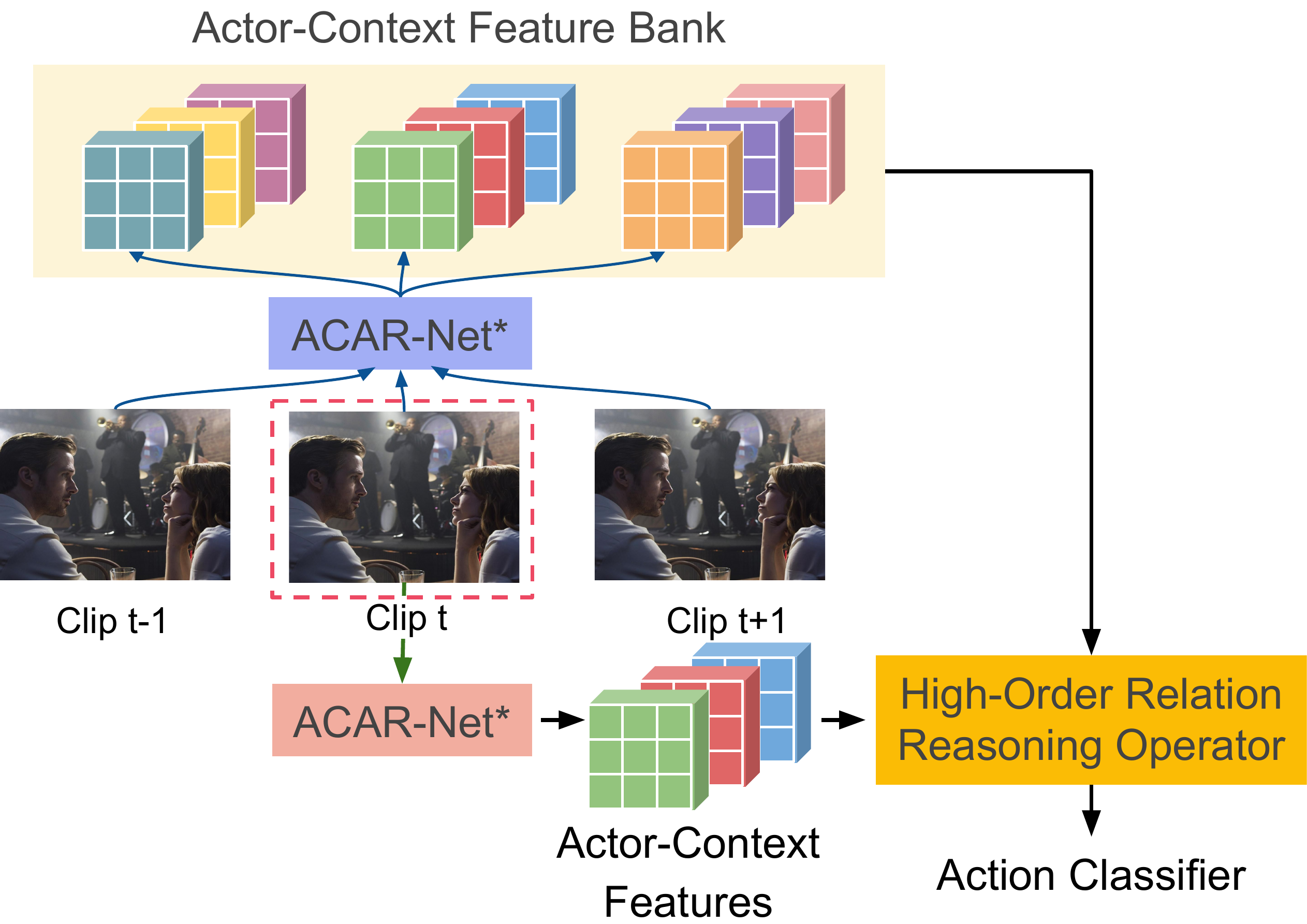}
\caption{Illustration of ACAR-Net equipped with Actor-Context Feature Bank, where ACAR-Net* refers to the first-order relation extraction part of our proposed module.
}
\label{fig:acfb}
\vspace{-5mm}
\end{figure}

In order to support actor-context-actor relation reasoning between actors and context at different time periods in a long video, we propose an Actor-Context Feature Bank (ACFB), in which we store contextual information from both past and future. This is inspired by the Long-term Feature Bank (LFB) proposed in \cite{wu2019long}. Yet instead of providing relational features for long-term higher-order reasoning, the previous LFB only stores actor features for facilitating first-order actor-actor interaction recognition.

As is illustrated in Fig.~\ref{fig:acfb}, clips are evenly sampled (every 1 second) from an input video, and the clips ($\sim$2s) could overlap with each other. We first train a separate ACAR-Net without any feature bank following the descriptions in Section \ref{sec:hro}.
First-order actor-context relation features $F^i$ of each actor in all clips of the entire video would be extracted by the separately pre-trained ACAR-Net and archived as the feature bank. To avoid confusion, we re-denote these acquired first-order features in the bank as $L^i$.
%

To train a new ACAR-Net with the support of the long-term actor-context feature bank to conduct high-order relation reasoning at some current time step $t$, we retrieve all $M$ archived actor-context relation features $\{L^i\}_{i=1}^M$ from the frames within a time window $[t-w, t+w]$. 
Actor-context-actor interactions between short-term features (encoding first-order interactions at current time $t$) and long-term ones from the archived bank can be computed as $\{H^i\}^{N}_{i=1}=\text{HR}^2\text{O}(\{F^i\}^{N}_{i=1},\{L^j\}^{{M}}_{j=1})$.
Note that, the $\text{HR}^2\text{O}$ is the same as before, but the self-attention mechanism is replaced with the attention between current and long-term actor-context relations, where query features $Q$ are still computed from short-term features $\{F^i\}^{N}_{i=1}$, but key and value features, $K$ and $V$, are calculated with the long-term archived features $\{{L}^i\}^{M}_{j=1}$, \ie,
\begin{equation}
    \begin{aligned}
    {Q^i}&=\mathrm{conv2D}(F^i),\\
    {K^j, V^j}&=\mathrm{conv2D}({L}^j).\\
    \end{aligned}
\end{equation}
Consequently, for any actor $i$ at current time $t$, our ACAR-Net is now capable of reasoning about its higher-relations with actors and context over a much longer time span, and thus better captures what is happening in the temporal context for achieving more accurate action localization.
\vspace{-2mm}

\section{Experiments on AVA}
\label{sec:Experiments}

AVA \cite{gu2018ava} is a video dataset for spatio-temporally localizing atomic visual actions. For AVA, box annotations and their corresponding action labels are provided on key frames of 430 15-minute videos with a temporal stride of 1 second. We use version 2.2 of AVA dataset by default.
In addition to the current AVA dataset, Kinetics-700~\cite{carreira2019short} videos with AVA~\cite{gu2018ava} style annotations are also introduced. The new AVA-Kinetics dataset~\cite{li2020ava} contains over 238k unique videos and more than 624k annotated frames. However, only a single frame is annotated for each video from Kinetics-700.
%
%
Following the guidelines of the benchmarks, we only evaluate 60 action classes with mean Average Precision (mAP) as the metric, using a frame-level IoU threshold of $0.5$. 



\subsection{Implementation Details}

{\flushleft \bf Person Detector.} For person detection on key frames, we use the human detection boxes from \cite{wu2019long}, which are generated by a Faster R-CNN~\cite{ren2015faster} with a ResNeXt-101-FPN~\cite{xie2017aggregated,lin2017feature} backbone. The model is pre-trained with Detectron~\cite{girshick2018detectron} on ImageNet~\cite{deng2009imagenet} as well as the COCO human keypoint images~\cite{lin2014microsoft}, and fine-tuned on the AVA dataset.
\vspace{-2mm}
{\flushleft \bf Backbone Network.} We use SlowFast networks~\cite{feichtenhofer2019slowfast} as the backbone in our localization framework and increase the spatial resolution of res$_5$ by $2\times$. We conduct ablation experiments using a SlowFast R-50 $8\times8$ instantiation (without non-local blocks). The inputs are 64-frame clips, where we sample $T=8$ frames with a temporal stride $\tau=8$ for the slow pathway, and $\alpha T (\alpha=4)$ frames for the fast pathway.  
The backbone is pre-trained on the Kinetics-400 dataset\footnote{The pre-trained SlowFast R-50 and SlowFast R-101+NL (in the following section) are downloaded from SlowFast's official repository.
}.

\vspace{-3mm}
\textbf{\flushleft Training and inference.} 
 In AVA, actions are grouped into 3 major categories: poses (e.g. stand, walk), human-object and human-human interactions. Given that poses are mutually exclusive and interactions are not, we use softmax for poses and sigmoid for interactions before binary cross-entropy loss for training.
We train all models end-to-end (except for
the feature bank part) using synchronous SGD with a batch size of 32 clips.
We train for 35k iterations with a base learning rate of 0.064, which is then decreased by a factor of 10 at iterations 33k and 34k. 
We perform linear warm-up~\cite{goyal2017accurate} during the first 6k iterations. We use a weight decay of $10^{-7}$ and Nesterov momentum of 0.9. 
We use both ground-truth boxes and predicted human boxes from \cite{wu2019long} for training. 
For inference, we scale the shorter side of input frames to 256 pixels and use detected person boxes with scores greater than 0.85 for final action classification.

\begin{table*}[!t]\vspace{-4mm}
\centering
%
%
\subfloat[\textbf{Relation Modeling}\label{tab:ablation:type}]{%
\tablestyle{4.8pt}{1.1}\begin{tabular}{@{}lx{20}@{}}
 & mAP \\
\shline
 {Baseline + STO} \cite{wu2019long} & 26.10\\
 {Baseline + ACRN \cite{sun2018actor} } & 26.71\\
 {Baseline + AIA} \cite{tang2020asynchronous} & 26.79\\
 {Baseline + HR$^2$O} & \textbf{27.83}\\
\end{tabular}}\hfill
\subfloat[\textbf{Component Analysis}\label{tab:ablation:structure}]{%
\tablestyle{4.8pt}{1.1}
\begin{tabular}{@{}lx{20}@{}}
& mAP \\
\shline
{Baseline} & 25.39\\
{Baseline + HR$^2$O} & {27.83}\\
{ACAR} & \textbf{28.84}\\
\multicolumn{2}{c}{~}\\
\end{tabular}}\hfill
\subfloat[\textbf{HR$^\mathbf{2}$O Design}\label{tab:ablation:function}]{%
\tablestyle{4.8pt}{1.1}
\begin{tabular}{@{}lx{25}@{}}
 & mAP\\
\shline
 {Avg \ \ \ \ \ } & 26.97 \\ 
 {RN} & 27.18 \\
 {NL} & \textbf{27.83}\\
 \multicolumn{2}{c}{~}\\
\end{tabular}}\hfill
%
%
%
\subfloat[\textbf{Relation Order}\label{tab:ablation:order}]{%
\tablestyle{4.8pt}{1.1}\begin{tabular}{@{}lx{20}@{}}
 & mAP \\
\shline
 {Actor First} & 27.62\\
 {Context First} & \textbf{27.83}\\
 \multicolumn{2}{c}{~}\\
\multicolumn{2}{c}{~}\\
\end{tabular}}\hfill
\subfloat[\textbf{Relation Depth}\label{tab:ablation:layer}]{%
\tablestyle{4.8pt}{1.1}
\begin{tabular}{@{}lx{20}@{}}
 & mAP\\
\shline
 {HR$^2$O-1L\ \ \ } & {27.63}\\
 {HR$^2$O-2L} & \textbf{27.83}\\
 {HR$^2$O-3L} & 27.25\\
\multicolumn{2}{c}{~}\\
\end{tabular}}\hfill
\subfloat[\textbf{Feature Bank}\label{tab:ablation:bank}]{%
\tablestyle{4.8pt}{1.1}\begin{tabular}{@{}lx{20}@{}}
 & mAP\\
\shline
 {HR$^2$O} & 27.83\\
 {HR$^2$O + LFB \cite{wu2019long}} & 27.75\\
 {HR$^2$O + ACFB} & \textbf{28.84}\\
\multicolumn{2}{c}{~}\\
\end{tabular}}

\vspace{2mm}
\caption{\textbf{Ablation study on AVA dataset}. The ``Baseline" of our framework only consists of the video backbone, actor detector and one-layer action classifier. HR$^2$O: High-order Relation Reasoning Operator. ACFB: Actor-Context Feature Bank. \label{tab:ablations}}
\vspace{-3mm}
\end{table*}

\subsection{Ablation Study}
We conduct ablation experiments to investigate the effect of different components in our framework on AVA v2.2. The baseline of our framework only consists of the video backbone (SlowFast R-50), the actor detector and the single-layer action classifier (denoted as ``Baseline'' in Table \ref{tab:ablations}).
\vspace{-3mm}
{\flushleft \bf Relation Modeling - Comparison.} In order to show the effectiveness of our actor-context-actor relation reasoning module, we compare against several previous approaches that leverage relation reasoning for action localization based on our baseline. Here we focus on validating the effect of relation modeling only, thus we disable long-term support in this study.
We adapt their reasoning modules such that all methods use the same baseline as our ACAR-Net in order to fairly compare only the impact of relation reasoning. We evaluate 
ACRN that focuses on learning actor-context relations; STO \cite{wu2019long} (a degraded version of LFB) that only captures actor interactions within the current short clip; AIA (w/o memory) \cite{tang2020asynchronous} that aggregates both actor-actor and actor-object interactions. As listed in Table~\ref{tab:ablation:type}, our proposed actor-context-actor relation modeling (``Baseline + HR$^2$O'' in Table~\ref{tab:ablation:type}) significantly improves over the compared methods. We observe that AIA with both actor and context relations performs better than ACRN and STO which only model one type of first-order relations, yet our method based on high-order relation modeling outperforms all compared methods by considerable margins. 

\begin{figure}[t]
\centering
\includegraphics[width=1\linewidth]{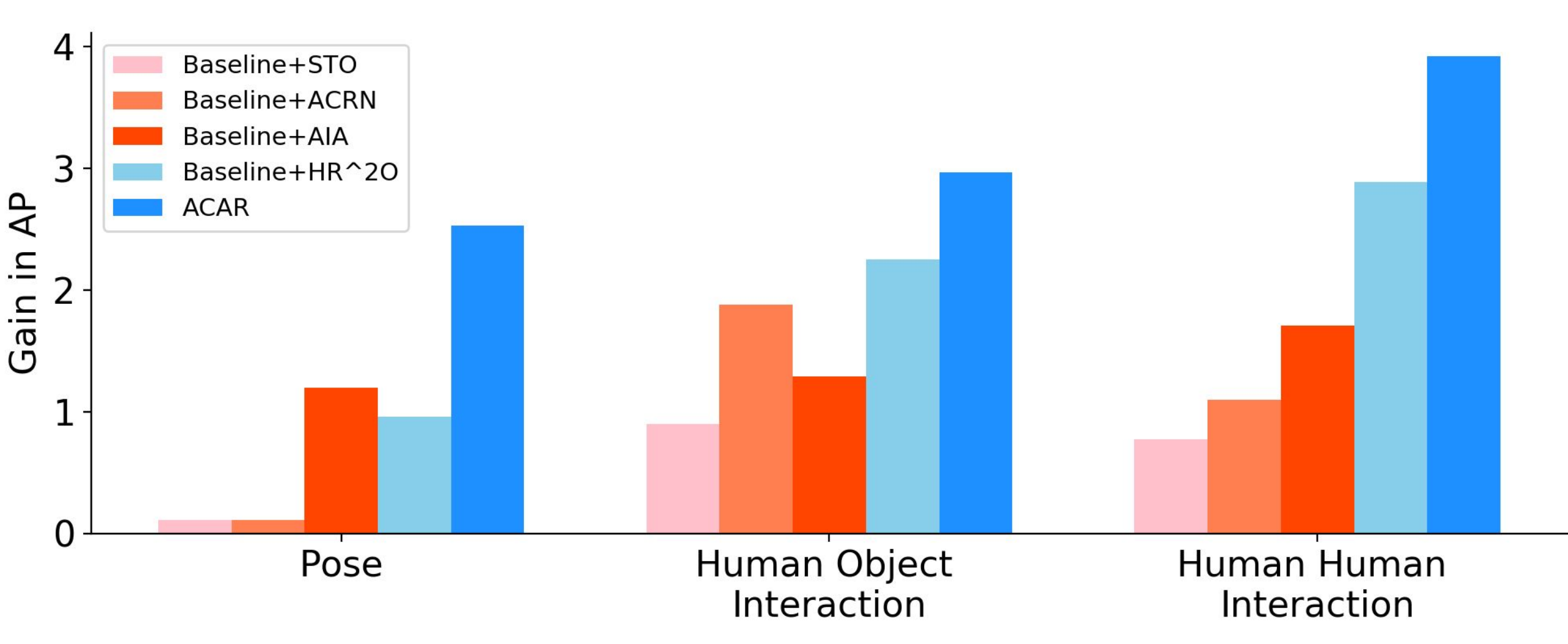}
\caption{\textbf{Gains of mAP on three major categories of the AVA dataset with respect to Baseline.} Our ACAR consistently outperforms other relation reasoning methods, and achieves larger performance gains on the two interaction categories. \label{fig:catgain}}

\end{figure}

We further break down the performances of different relation reasoning modules into three major categories of the AVA dataset, which are poses (\textit{e.g.} stand, sit, walk), human-object interactions (\textit{e.g.} read, eat, drive) and human-human interactions (\textit{e.g.} talk, listen, hug). Fig.~\ref{fig:catgain} compares the gains of different approaches with respect to the baseline in terms of mAP on these major categories. We can see that our HR$^2$O gives more performance boosts 
on two interaction categories compared to the pose category, which is consistent with our motivation to model indirect relations between actors and context. Once equipped with ACFB, our framework can further improve on the pose category as well.

\begin{figure}[t]
\centering
\includegraphics[width=1\linewidth]{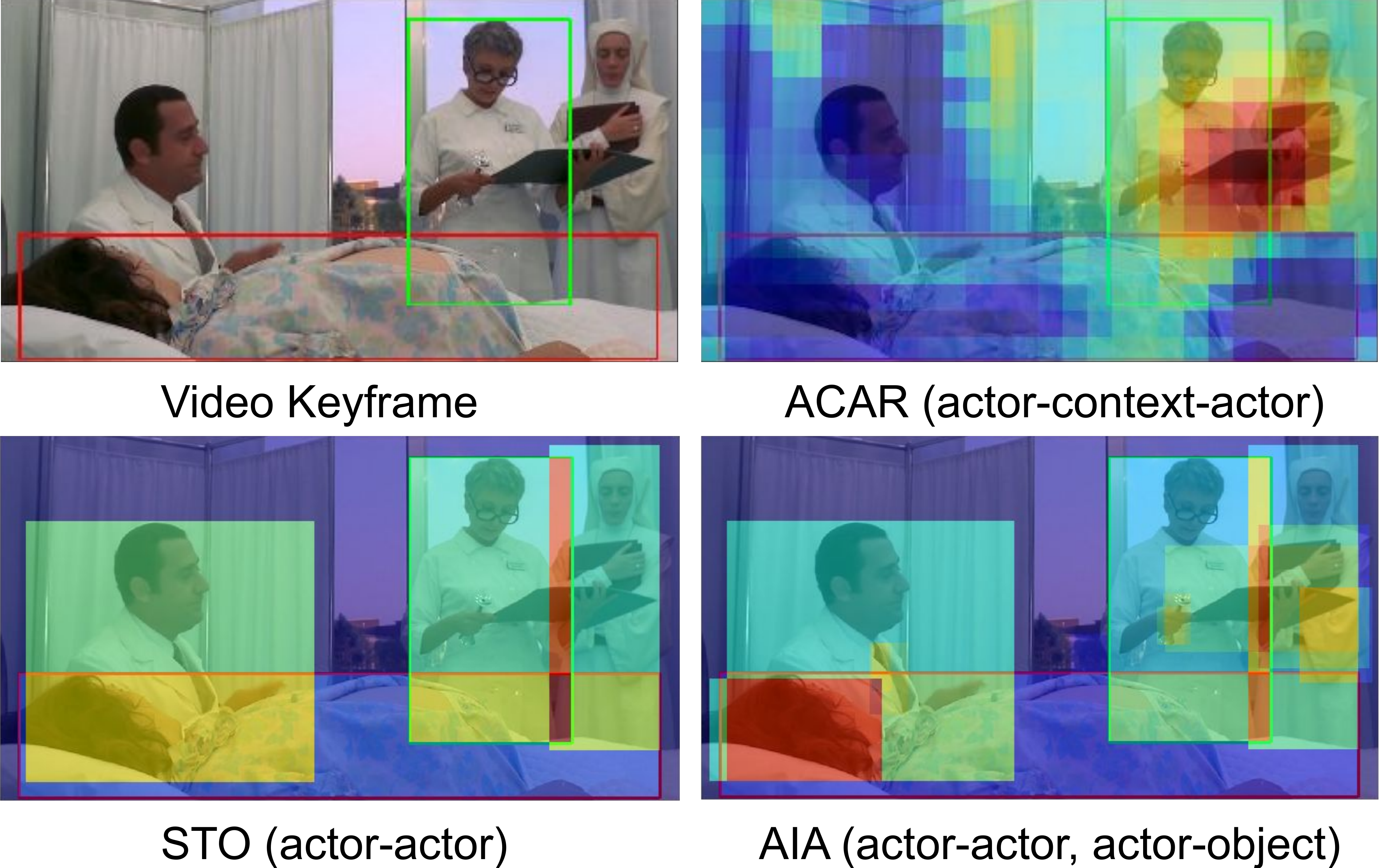}
\caption{\textbf{Comparison of attention maps from different approaches of relation modeling for action detection.}
Our method is able to attend the contextual region (some document) that relates the \textcolor{red}{actor of interest} marked in red (performing ``listen to'') and the \textcolor{green}{supporting actors} in the green box (performing ``read''), while other methods fail to achieve similar effects.  \label{fig:compare_attmap}}
\vspace{-4mm}
\end{figure}

Finally, we contrast our ACAR with existing relation reasoning approaches in AVA. We visualize attention maps from different reasoning modules over an example key frame in Fig.~\ref{fig:compare_attmap}. Without needing object proposals, ACAR is capable of localizing free-form context regions for indirectly establishing relations between two actors (the actor of interest is listening to the supporting actor reading a report). In comparison, the attention weights of STO as well as AIA are distributed more diversely and do not have a clear focus point. Note that we do not show the attention map of ACRN since it assigns equal weights to all context regions. 
{\flushleft \bf Component Analysis.} To validate our design, we first ablate the impacts of different components of our ACAR as shown in Table \ref{tab:ablation:structure}. We can observe that both HR$^2$O and ACFB lead to significant performance gains over baseline. 
%
{\flushleft \bf HR$^\mathbf{2}$O Design.} We test different instantiations of the High-order Relation Reasoning Operator on top of our baseline in Table \ref{tab:ablation:function}. 
Our modified non-local (denoted as ``NL'') mechanism works better than simply designing $\text{HR}^2\text{O}$ as an average function (denoted as ``Avg''), \ie $H^i = \frac{1}{N} \sum_{i} F^i,$. In addition, the instantiation with relation network (RN) described in Section \ref{sec:hro} also works 
alright. Nonetheless, the modified non-local attention is computationally more efficient than RN with feature triplets and has better performance.
%
{\flushleft \bf Relation Ordering.} There are two possible orders for reasoning actor-context-actor relations: 1) aggregating actor-actor relations first, or 2) encoding actor-context relations first. Note that our ACAR-Net adopts the latter one. We implement the former order by performing self-attention between actor features with the modified non-local attention before incorporating context features in our baseline. The results in Table~\ref{tab:ablation:order} validate that context information should be aggregated earlier for better relation reasoning.
%
{\flushleft \bf HR$^\mathbf{2}$O Depth.} In Table~\ref{tab:ablation:layer}, we observe that stacking two modified non-local blocks in HR$^2$O has higher mAP than the one-layer version, yet adding one more non-local block produces worse performance, possibly due to overfitting. We therefore adopt two non-local blocks as the default setting.
{\flushleft \bf Actor-Context Feature Bank.} In this set of experiments, we validate the effectiveness of the proposed ACFB. We set the ``window size'' $2w+1$ to 21s due to memory limitations, and longer temporal support is expected to perform better \cite{wu2019long}. 
As presented in Table~\ref{tab:ablation:bank}, adding long-term support with ACFB significantly improves the baseline (HR$^2$O's 27.83 $\rightarrow$\ HR$^2$O + ACFB's \textbf{28.84}).
We also test replacing the ACFB in our framework with the long-term feature bank (LFB)~\cite{wu2019long} (denoted as ``HR$^2$O + LFB''). However, LFB even fails to match the baseline performance. This drop might be because LFB encodes only ``zeroth-order'' actor features, which cannot provide enough relational information from neighboring frames for assisting interaction recognition.


\begin{table}[!t]
\centering
\tablestyle{4.8pt}{1.1}\begin{tabular}{@{}l|x{23}|x{37}@{}}
 model & inputs & val mAP\\
\shline
 {ACRN, S3D} \cite{sun2018actor} & V+F & 17.4 \\
 {Zhang \etal} \cite{zhang2019structured}, I3D & V & 22.2 \\
 {Action TX, I3D} \cite{girdhar2019video} & V & 25.0 \\
 {LFB}, R-50+NL \cite{wu2019long} & V & 25.8 \\
 {LFB}, R-101+NL \cite{wu2019long} & V & 27.4 \\
 {SlowFast, R-50, $8\times8$} \cite{feichtenhofer2019slowfast} & V & 24.8 \\
 {SlowFast, R-101, $8\times8$} \cite{feichtenhofer2019slowfast} & V & 26.3 \\
 \hline
 \textbf{Ours}, R-50, $8\times8$ & V & 28.3 \\
 \textbf{Ours}, R-101, $8\times8$ & V & {\bf 30.0} \\
\end{tabular}
\vspace{2mm}
\caption{\textbf{Comparison with state-of-the-arts on AVA v2.1.} 
All models are pre-trained on Kinetics-400. V and F refer to visual frames and optical flow respectively. \label{tab:sota_v2.1}}
\end{table}

\begin{table}[!t]
\centering
\tablestyle{4.8pt}{1.1}\begin{tabular}{@{}l|x{51}|x{37}@{}}
 model & pre-train & val mAP \\
\shline
 {SlowFast, R-101+NL} \cite{feichtenhofer2019slowfast} & Kinetics-600 & 29.0 \\
 {AIA, R-101+NL} \cite{tang2020asynchronous} & Kinetics-700 & 32.3 \\
 \hline
 Ours, R-101+NL & Kinetics-600 & 31.4 \\
 \textbf{Ours}, R-101 & Kinetics-700 & {\bf 33.3} \\
\end{tabular}
\vspace{2mm}
\caption{\textbf{Comparison with state-of-the-arts on AVA 2.2.} We do not conduct testing with multiple scales and flips. All models use $T\times\tau=8\times8$. \label{tab:sota_v2.2}}
\end{table}

\begin{table}[!t]
\centering
\tablestyle{4.8pt}{1.1}\begin{tabular}{@{}l|x{37}|x{37}@{}}
 model & val mAP & test mAP\\
\shline
 {AIA++, ensemble \cite{xia2020report}}& - & 32.91 \\
 {MSF, ensemble \cite{Zhu2020report}}& - & 31.88 \\
\hline
 {SlowFast, R-101, $8\times8$ (our impl.)} & 32.98 & - \\
 {Ours, R-101, $8\times8$} & 35.84 & - \\ 
 {Ours++, R-101, $8\times8$} & 36.36 & - \\
 {Ours++, ensemble}  & \textbf{40.49} & \textbf{39.62}
\end{tabular}
\vspace{2mm}
\caption{\textbf{AVA-Kinetics results}. ``++'' refers to inference with 3 scales and horizontal flips. Models submitted to the test server are trained on both training and validation sets.
\label{tab:sota_ak}}
\vspace{-3mm}
\end{table}


\subsection{Comparison with State-of-the-arts on AVA}

We compare our ACAR-Net with state-of-the-art methods on the validation set of both AVA v2.1 (Table~\ref{tab:sota_v2.1}) and v2.2 (Table~\ref{tab:sota_v2.2}). Note that we also provide results with more advanced video backbones, \ie two SlowFast R-101 instantiations (with / without NL). 
On AVA v2.1, our framework achieves {30.0 mAP} and outperforms all prior results with pre-trained Kinetics-400 backbone. On AVA v2.2, our ACAR-Net reaches {33.3 mAP} with only single-scale testing, establishing a new state-of-the-art. Note that our method surpasses AIA \cite{tang2020asynchronous} with only $1/3$ of temporal support. The results indicate that with proper modeling of higher-order relations, our approach extracts more informative cues from the context. 

\begin{figure*}[t]
\centering
\includegraphics[width=1\linewidth]{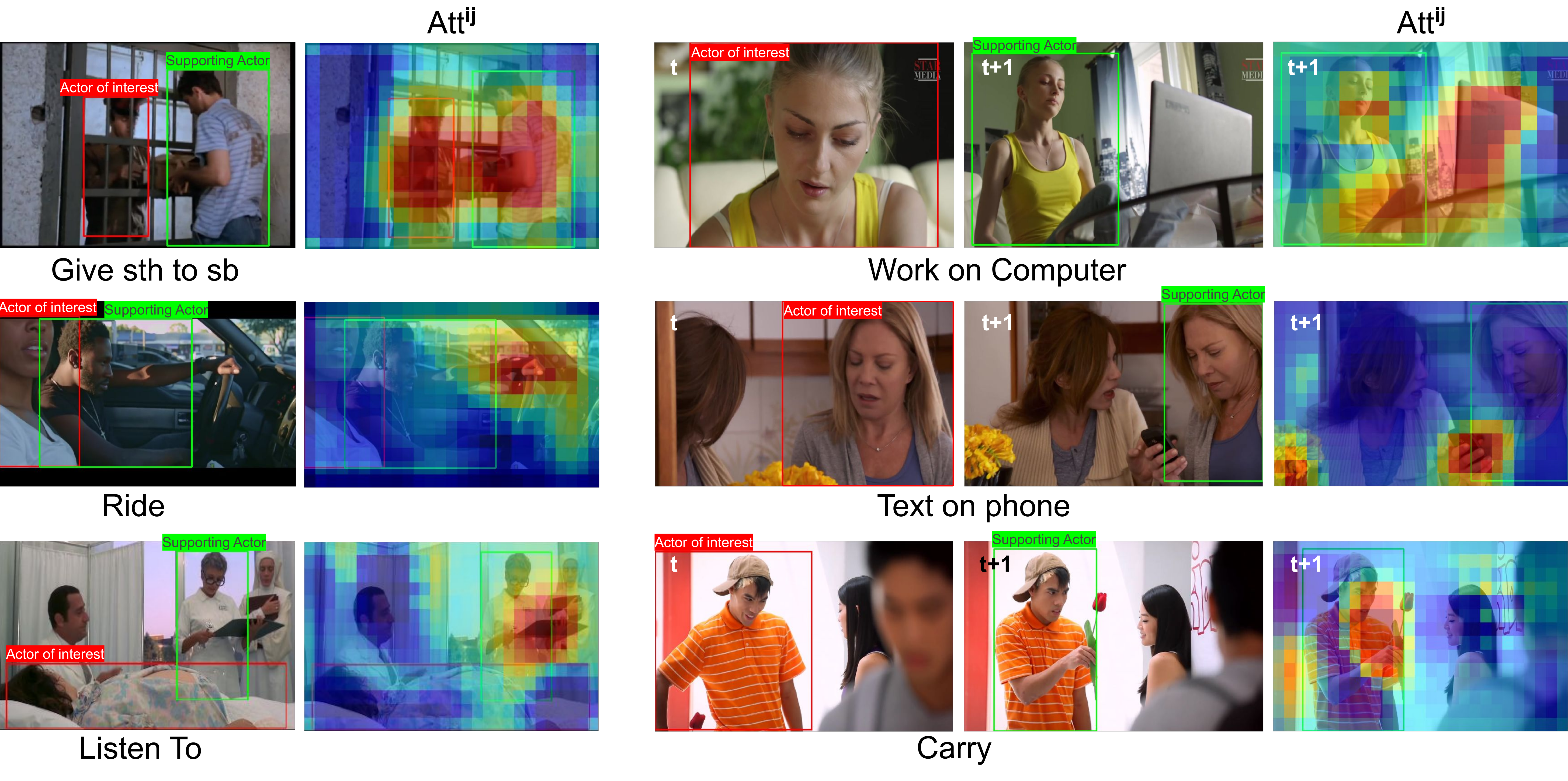}
\caption{\textbf{Visualization of actor-context-actor attention maps on AVA.} \textcolor{red}{Actors of interest} are marked in red and \textcolor{green}{supporting actors} in green. Heat maps illustrate the context regions' attention weights $Att^{i,j}$ from actor-context-actor relation reasoning. We observe that our model has learned to attend to useful relations between actors and context, and the context serves as the bridge for connecting actors.
\label{fig:attn}}
\vspace{-4mm}
\end{figure*}


We present our results on AVA-Kinetics in Table~\ref{tab:sota_ak}. Our baseline is already highly competitive ($\sim$33 mAP). Yet integrating our ACAR modeling still leads to a significant gain of +2.86 mAP. This demonstrates that performance enhancement brought by high-order relation modeling can generalize to this new dataset.
With an ensemble of models, we achieve \textbf{39.62 mAP} on the test set, ranking first in the AVA-Kinetics task of ActivityNet Challenge 2020 and outperforming other entries by a large margin (+6.71 mAP). More details on our winning solution are provided in the technical report \cite{chen20201st}.

\subsection{Qualitative Results} 

Our proposed ACAR operates fully convolutionally on top of spatio-temporal features, and this allows us to visualize the actor-context-actor relation maps $\{Att^{i,j}\}$ generated by our High-order Relation Reasoning Operator. As shown in Fig.~\ref{fig:attn}, the first two columns include the key frame as well as the corresponding relation map from the same clip, and the last three columns show the relation map denoting interactions with actors and context from a neighboring clip.
We can observe that the attended regions usually include the actor of interest, supporting actors' body parts (\ie head, hands and arm) and objects being in interaction with the actors. Take the first example on the left as an example. The green supporting actor $A^j$ is taking a package from the red actor of interest $A^i$. Such information is well encoded by our ACAR-Net in the form of actor-context-actor relations: packages, hands and arms of both actors are highlighted.

\section{Experiments on UCF101-24}
\label{sec:Experiments_UCF}
UCF101-24 is a subset of UCF101~\cite{soomro2012ucf101} that contains spatio-temporal annotations for 3,207 videos on 24 action classes. Following the evaluation settings of previous methods \cite{kalogeiton2017action, yang2019step}. We experiment on the first split and report frame-mAP with an IoU threshold of 0.5.
\vspace{-2mm}
{\flushleft \bf Implementation Details.} We also use SlowFast R-50 pre-trained on Kinetics-400 as the video backbone, and adopt the person detector from \cite{kopuklu2019you}. The temporal sampling for the slow pathway is changed to $8\times4$ and the fast pathway takes as input 32 continuous frames.

For training, we train all the models end-to-end for 5.4k iterations with a base learning rate of 0.002, which is then decreased by a factor of 10 at iterations 4.9k and 5.1k.
We perform linear warm-up during the first quarter of the training schedule. 
We only use ground-truth boxes for training, and use all boxes given by the detector for inference. Other hyper-parameters are similar to the experiments on AVA.
\vspace{-2mm}
{\flushleft \bf Results.} As shown in Table~\ref{tab:ucf}, ACAR surpasses the strong baseline with a considerable margin, which again indicates the importance of high-order relation reasoning.

\begin{table}[t]
\centering
\tablestyle{4.8pt}{1.1}\begin{tabular}{@{}l|x{23}|x{31}@{}}
 model & inputs & mAP\\
\shline
 {T-CNN} \cite{hou2017tube} & V &67.3 \\
 {ACT} \cite{kalogeiton2017action} & V &69.5 \\
 {STEP, I3D} \cite{yang2019step} & V+F &75.0 \\
 {I3D} \cite{gu2018ava} & V+F & 76.3 \\
 {Zhang \textit{et al.}} \cite{zhang2019structured}, I3D & V & 77.9\\
 {S3D-G} \cite{xie2018rethinking} & V+F &78.8\\
 {AIA, R-50} \cite{tang2020asynchronous} & V &78.8\\
 \hline
 SlowFast R-50, $8\times4$ (ours) & V &\underline{82.4} \\
 \textbf{Ours w/o ACFB}, R50, $8\times4$ & V &{\bf 84.3} \\
\end{tabular}
\vspace{3mm}
\caption{\textbf{Comparison with previous works on UCF101-24.} We evaluate frame-mAP on split 1. V and F refer to visual frames and optical flow respectively.\label{tab:ucf}}
\vspace{-5mm}
\end{table}

\section{Conclusion}
\label{sec:Conclusion}

Given the high complexity of realistic scenes encountered in the spatio-temporal action localization task which involve multiple actors and a large variety of contextual objects, we observe the demand for a more sophisticated form of relation reasoning than current ones which often miss important hints for recognizing actions. 
Therefore, we propose Actor-Context-Actor Relation Network for explicitly modeling higher-order relations between actors based on their interactions with the context.
Extensive experiments on the action detection task show our ACAR-Net outperforms existing methods that leverage relation reasoning, and achieves state-of-the-art results on several challenging benchmarks of spatio-temporal action localization.

\noindent {\bf Acknowledgements.} We thank Charlie W., Jiajun T. and Daixin W. for helpful discussions. This work is supported in part by the General Research Fund through the Research Grants Council of Hong Kong under Grants (Nos. 14208417, 14207319, 14202217, 14203118, 14208619), in part by Research Impact Fund Grant No. R5001-18, in part by CUHK Strategic Fund.

{\small
\bibliographystyle{ieee_fullname}
\bibliography{egbib}
}
\end{document}